\documentclass[10pt,twocolumn,letterpaper]{article}

\usepackage{iccv}
\usepackage{bm}
\usepackage{times}
\usepackage{epsfig}
\usepackage{graphicx}
\usepackage{amsmath}
\usepackage{amssymb}
\usepackage{mathrsfs}
\usepackage[table,xcdraw]{xcolor}

\usepackage[pagebackref=true,breaklinks=true,letterpaper=true,colorlinks,bookmarks=false]{hyperref}

\iccvfinalcopy 

\def\iccvPaperID{2554} 

\ificcvfinal\fi
\begin{document}

\title{Discriminative convolutional Fisher vector network for action recognition}

\author{Petar Palasek, Ioannis Patras\\
School of Electrical Engineering and Computer Science\\
Queen Mary University of London\\
London E1 4NS, United Kingdom\\
{\tt\small p.palasek@qmul.ac.uk},  {\tt\small i.patras@qmul.ac.uk}
}

\maketitle

\begin{abstract}
In this work we propose a novel neural network architecture for the problem of human action recognition in videos. The proposed architecture expresses the processing steps of classical Fisher vector approaches, that is dimensionality reduction by principal component analysis (PCA) projection, Gaussian mixture model (GMM) and Fisher vector descriptor extraction, as network layers. By contrast to other methods where these steps are performed consecutively and the corresponding parameters are learned in an unsupervised manner, having them defined as a single neural network allows us to refine the whole model discriminatively in an end to end fashion. 
Furthermore, we show that the proposed architecture can be used as a replacement for the fully connected layers in popular convolutional networks achieving a comparable classification performance, or even significantly surpassing the performance of similar architectures while reducing the total number of trainable parameters by a factor of 5. We show that our method achieves significant improvements in comparison to the classical chain.
\end{abstract}

\section{Introduction}

With the amounts of video data available online growing at high rates, the need for automatic video analysis is becoming more and more pressing. Being able to automatically recognize what the content of a given video, or, more narrowly recognize the actions that are depicted in it, is not only useful for organizing huge video datasets, but also something that could help improve systems for video surveillance, human-computer interaction systems and assistance systems.

Due to their strong performance on the problem of image recognition \cite{perronnin2010improving}, Fisher vector (FV) descriptors \cite{sanchez2013image} have also been applied for the problem of action recognition \cite{wang2011action, wang2013action} where they achieved state of the art results and remained one of the dominant approaches to this day.

The main idea behind FVs is to encode a set of local descriptors extracted from a sample (\ie an image or a video) as a vector of deviations from the parameters of a generative model (usually a Gaussian mixture model) fitted to the descriptors extracted on the training set. Although FVs are good global descriptors on their own, there are shortcomings in the way they are extracted. Namely, the GMM used for encoding is learnt in an unsupervised way 
without receiving any additional information about the task at hand. This results in descriptors that are not tailored for a discriminative task as the GMM also learns to model the intra-class variations of the training data something that is not relevant for classification problems.

Recently, methods that are based on multi-layered convolutional \cite{lecun1998gradient} neural networks (CNNs) surpassed the Fisher vector descriptors' performance in recognition problems and reached the new state of the art in image classification. The seminal work described in \cite{krizhevsky2012imagenet} has shown the power of models that can be  trained end to end in a supervised way on large amounts of labeled data using backpropagation. This is one of the works that helped regain popularity of neural networks and start the deep learning revolution. Many works on deep networks have been published since then, not only for the task of image recognition \cite{chatfield2014return, simonyan2014very}, but for the problem of action recognition \cite{feichtenhofer2016convolutional, ji20133d,karpathy2014large,simonyan2014two,tran2015learning} as well.

However, the state of the art deep learning architectures usually contain a large number of layers with a huge number of trainable parameters making them difficult to optimize without suitable hardware infrastructure (\ie big clusters or machines with multiple GPUs have become a necessity nowadays).

Recently several works that try to combine the power of Fisher vector representations and neural network approaches have been published. This includes deep Fisher networks from \cite{simonyan2013deep} used for large-scale image classification, stacked Fisher vectors \cite{peng2014action} used for action recognition, deep Fisher kernels \cite{sydorov2014deep} and the hybrid classification architecture \cite{perronnin2015fisher} also applied on image classification problems. Even though the works listed above explore adding supervision at different steps of the standard Fisher vector descriptor pipeline, none of them have tried to refine the local feature extraction, Fisher vector encoding and the classification steps jointly. 

In this paper we describe a method that expresses all the different steps of the action recognition using FV, as layers in a neural network. The layers are initialized by unsupervised training in a layer by layer manner, and are subsequently refined in an end-to-end training. The proposed architecture results in spatio-temporal descriptors at intermediate levels of the architecture, calculated by local aggregation in a spatio-temporal structure of frame-level descriptors. 
More specifically, the main contributions of this work are the following:

\begin{itemize}

  \item We describe a novel neural network architecture for action recognition which includes two new types of layers; the Gaussian mixture model layer and the Fisher vector descriptor layer. Combining these layers with other standard ones into a single deep neural network gives us a way of jointly finetuning the parameters of the whole architecture with respect to a chosen discriminative cost, using the standard backpropagation algorithm. We show that adding supervision at every stage of the network improves the discriminative power of the extracted Fisher vector descriptor compared to the standard version of the descriptor which is extracted in an unsupervised manner.
      
 \item Analogous to convolutional neural networks, where the same operation is applied at different locations of the input tensor, our network offers a natural way of extracting the Fisher vector descriptors densely from a given input video, both in space and in time. This also allows us to easily extract the descriptor only from selected parts of the video, providing a straightforward way of implementing other architectures, such as spatial pyramids.
 
 \item  We show that the proposed architecture can be used as a replacement for the fully connected layers in popular convolutional networks such as the VGG-16 network, achieving a comparable classification performance while reducing the total number of trainable parameters by a factor of 5.

\end{itemize}

\section{Related work}

Fisher vector descriptors were firstly introduced for solving the problem of image classification in \cite{perronnin2007fisher}. Essentially, the idea is to represent an image using a global descriptor which describes how the parameters of a generative model should change in order to better model the distribution of local features in images, based on a set of local features extracted from a given image. The theory and practice of using Fisher vectors for the task of image classification is described in \cite{sanchez2013image}.

The first work that applied Fisher vector descriptors for the problem of action recognition in videos used HOG, HOF and MBH features \cite{wang2009evaluation} extracted along dense trajectories as local features \cite{wang2011action}. The trajectories are extracted by defining a dense grid of points which are then tracked using optical flow that was estimated offline, this way including motion information in the pipeline. By encoding the extracted trajectory features with the Fisher vector descriptor, this approach and the improved version of \cite{wang2013action} achieved state of the art results for the action recognition problem.

Following the growing popularity of deep neural networks, several works were published on using neural network based approaches for the problem of action recognition. In \cite{ji20133d} a 3D extension of the standard 2D convolutional neural networks (CNNs) was introduced, where information from both the space and the time dimension are included by performing 3D convolution. A more recent work described in \cite{tran2015learning} also applied 3D CNNs, but with a much deeper network architecture.  The work of \cite{karpathy2014large} examines different kinds of extending CNNs into the time domain by fusing features extracted from stacks of frames in order to include motion information. Transfer learning is also applied in order to prevent overfitting to small video datasets. Motion information was included in the work of \cite{simonyan2014two} in an explicit way by providing dense optical flow at the input of the network. More specifically, two streams of a network are employed; one performing classification based on static video frames and the other based on the optical flow. Different ways of fusing the spatial and the temporal streams of such networks are studied in \cite{feichtenhofer2016convolutional}. The work of \cite{yue2015beyond} considers different ways of aggregating strong CNN image features over long periods of time, including feature pooling and using recurrent neural networks. In \cite{palasek2016marmi} CNN features are extracted from random subvolumes of a video and encoded using the FV descriptor in order to arrive at a representation suitable for video classification.

The work of \cite{simonyan2013deep} combines ideas from the area of neural networks with the Fisher vector descriptor by forming a deep Fisher network in which two Fisher vector layers are stacked. The network is discriminatively trained for the problem of image classification, however the features at the input layer are fixed, manually-designed features. Stacked Fisher vectors are also applied for action recognition in \cite{peng2014action}, where the first layer encodes the improved dense trajectories from \cite{wang2013action}. After discriminative dimensionality reduction a second Fisher vector encoding is done. The combination of the FV and the stacked FV showed to be beneficial. The work in \cite{perronnin2015fisher} treats the Fisher vector descriptor as an unsupervised layer  followed by a number of fully connected layers. that can be trained with backpropagation. End to end training of a Fisher kernel SVM viewed as a deep network is done in \cite{sydorov2014deep} for the problem of image classification. This method uses manually-designed features at the input layer and requires retraining of the SVM on the whole training set at each step of the training.

\section{Proposed architecture}

\begin{figure*}[t]
\begin{center}
\includegraphics[width=0.95\linewidth]{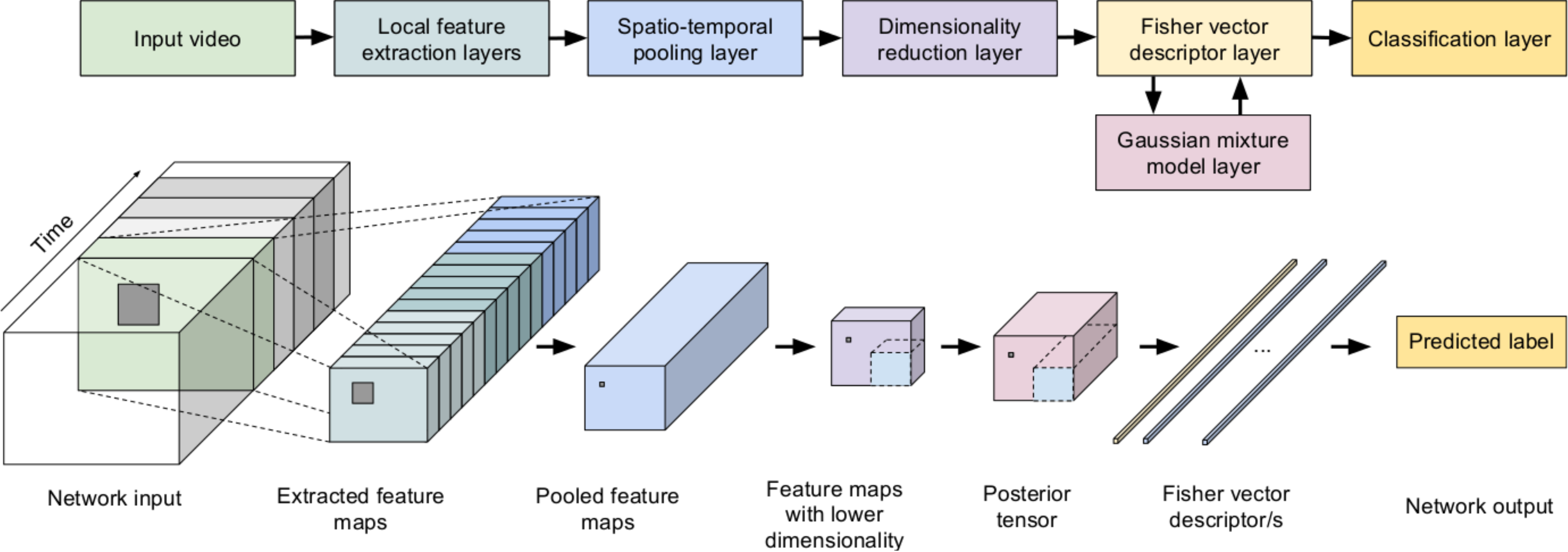}

\end{center}
   \caption{An illustration of the proposed architecture. The input to the network is a stack of $t$ static frames (marked in green in the left of the figure) from a video which are passed through the feature extraction layers resulting in $t$ feature maps of size $F_h \times F_w \times d$ where $d$ is the number of channels. The feature maps are then pooled temporally and spatially which results in new feature maps of size $F_h' \times F_w' \times D$ whose dimensionality is reduced in the dimensionality reduction layer to $F_h'\times F_w' \times n_c$. The Fisher vector descriptor layer passes these feature maps to the GMM layer which gives a tensor of posteriors at its output. Using the posteriors and the input feature maps, the FV layer outputs the FV descriptor of the $t$ frames of the video. Note that we can use different crops of the posterior tensor in order to calculate the FV descriptor of only a part of the network input (\eg using the subtensor marked in blue in the posterior tensor would correspond to calculating the FV descriptor for only the bottom right corner of the input video). Finally the FV descriptors are normalized and fed into the classification layer where their scores are averaged and used to predict the label for the given input. In order to predict a label for the whole video of length $L$, we slide the network along the time axis with a stride of $\delta_T$ frames, updating the FV descriptor/s on the way. The classification step is the same as when predicting the label of a stack of $t$ frames.}

\label{fig:architecture}
\end{figure*}

In this section we describe the proposed architecture  - an illustration is given in Figure \ref{fig:architecture}. We start by describing each of the used layers in detail and give the final overview of the whole architecture in Subsection \ref{section:final}.

The architecture can be divided into six parts; the local feature extraction layers, the spatio-temporal pooling layer, the dimensionality reduction layer, the Gaussian mixture layer, the Fisher vector descriptor layer and the classification layer. Their descriptions follow in Subsections \ref{section:extraction}, \ref{section:pooling}, \ref{section:reduction}, \ref{section:gmm}, \ref{section:fisher} and \ref{section:class}.


\subsection{Local feature extraction layers}
\label{section:extraction}
To do the first step of local feature extraction in our architecture, we tried using two different networks. We fed static video frames into a small network consisting of a single convolutional layer followed by a pooling layer. This part of the network was pretrained on static video frames using a convolutional restricted Boltzmann machine \cite{lee2009convolutional} as described in \cite{palasek2016marmi}, with the difference that we used local contrast normalization as a preprocessing step. To show that the feature extraction layers can be replaced by any larger and more complex network, we also used the VGG-16 \cite{simonyan2014very} network pretrained on the ImageNet dataset. Given $L$ consecutive images, the output of the feature extraction layers is $L$ feature maps of size $F_h \times F_w \times d$, where $F_h$ denotes its height, $F_w$ the width and $d$ the number of its channels.

\subsection{Spatio-temporal pooling layer}
\label{section:pooling}

\begin{figure*}[t]
\begin{center}
\includegraphics[width=0.62\linewidth]{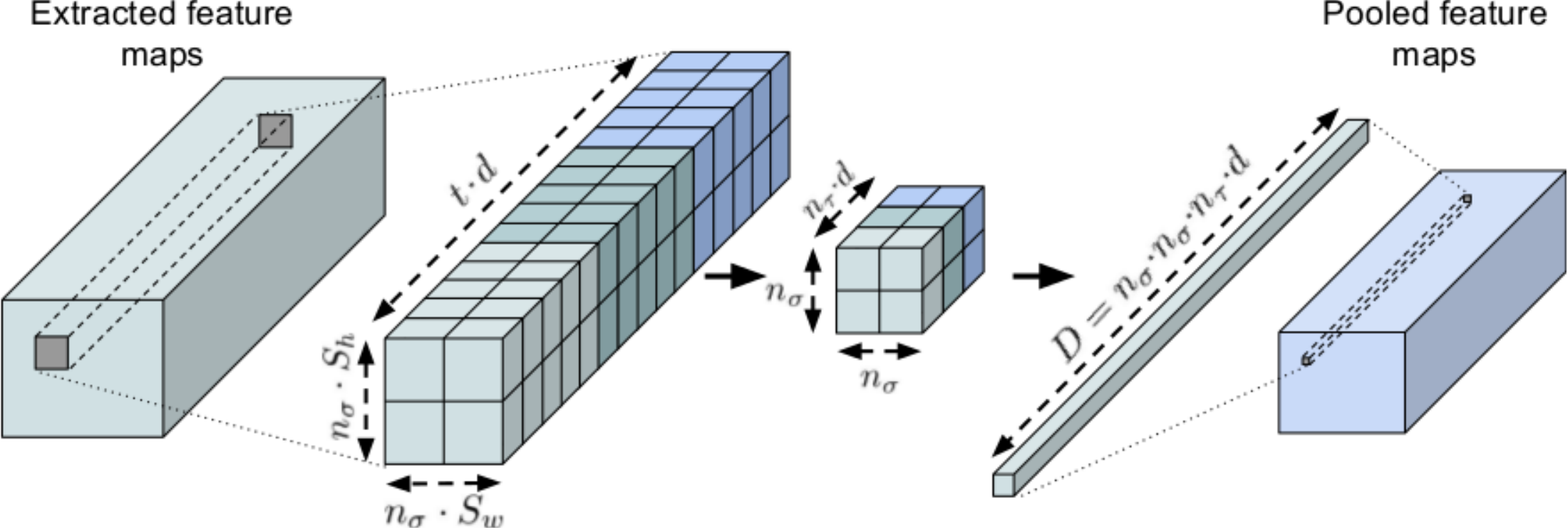}

\end{center}
   \caption{An illustration of how the spatio-temporal layer performs pooling on the extracted feature maps. Subtensors of size $n_\sigma \cdot S_h \times n_\sigma \cdot S_w \times t \cdot d$ are pooled into tensors of size  $n_\sigma \times n_\sigma \times n_\tau \cdot d$ and resized into vectors of dimensionality $D$. This is repeated for all subtensors in the extracted feature maps, sliding horizontally and vertically with a stride of $\delta_S$.}

\label{fig:pooling}
\end{figure*}
In order to include motion information from the input video, we want to combine feature maps extracted from multiple static frames into a more powerful representation. To do so, we follow the work of \cite{wang2011action} where a spatio-temporal volume of features extracted from $t$ frames is divided into $n_\sigma \times n_\sigma \times n_\tau$ subvolumes of size $S_h \times S_w \times t / n_\tau$ and then pooled temporally and spatially using mean pooling. The resulting representation is then resized into a vector of dimensionality $D=n_\sigma \cdot n_\sigma \cdot n_\tau \cdot d$, where $d$ is the dimensionality of the local features extracted from the previous layer. This vector is then used as input to the following layer.
In the case that the tensor at the input of this layer has bigger spatial dimensions than $n_\sigma \cdot S_h \times n_\sigma \cdot S_w$, the described pooling procedure is repeated for each $n_\sigma \cdot S_h \times n_\sigma \cdot S_w \times t \cdot d$ subtensor, moving through the tensor with a spatial stride $\delta_S$ as shown in Figure \ref{fig:pooling}. Later we will describe how we deal with longer time periods. Given $t$ feature maps of size $F_h\times F_w \times d$ at input, the output of this layer is a tensor of size $F_h' \times F_w' \times D$, where $F_h'=(F_h - S_h) / \delta_S + 1$ and $F_w'=(F_w - S_w) / \delta_S + 1$.


\subsection{Dimensionality reduction layer}
\label{section:reduction}
In the standard Fisher vector pipeline the locally extracted features are decorrelated and their dimensionality is reduced by performing PCA. Assuming that the mean of the data $\mu_x$ and the principal axes $P$ were found offline, the mapping from the original data to a lower-dimensional space can be written as:
\begin{equation}
\bm{x}_t'=(\bm{x}_t-\bm{\mu}_{x})\bm{P}'.
\end{equation}
The dimensionality of matrix $\bm{P}$ is $n_c \times D$, where $n_c$ is the number of components and $D$ is the dimensionality of original data. 
Note that we do not put any constraints on the matrix $\bm{P}$, so after backpropagating through the layer and updating its parameters the projection applied on the input data is not guaranteed to be orthogonal. Given a tensor of size $F_h' \times F_w' \times D$ at input, the output of this layer is a tensor of size $F_h' \times F_w' \times n_c$.

\subsection{Gaussian mixture model layer}
\label{section:gmm}
A Gaussian mixture model is defined as a weighted sum of K components \cite{sanchez2013image}:
\begin{equation}
u_\lambda(\bm{x}) = \sum_{k=1}^{K}w_k u_k(\bm{x}),
\end{equation}
where $w_k$ is a component weight and $u_k(\bm{x})$ is a probability density function of the Gaussian distribution:
\begin{equation}
u_k(\bm{x}) = \frac{1}{(2\pi)^{\frac{D}{2}}\lvert\bm{\Sigma}_k\rvert^{\frac{1}{2}}}\exp\left( -\frac{1}{2}(\bm{x}-\bm{\mu}_k)' \bm{\Sigma}_k^{-1} (\bm{x} - \bm{\mu}_k) \right).
\end{equation}
Every GMM can be described by the parameter set $\lambda = \{w_k, \bm{\mu}_k, \bm{\Sigma}_k, k = 1, ..., K \}$, where $w_k$ is the $k$-th component weight, $\bm{\mu}_k$ is its mean vector and $\bm{\Sigma}_k$ its covariance matrix. The mixture coefficients $\{w_k\}$ are constrained to be positive and to sum to one, that is $w_k\geq 0$ and $\sum_{k=1}^{K}{w_k}=1$ which can be easily enforced by using internal weights $\alpha_k$ and defining:
\begin{equation}
w_k=\frac{\exp \left(\alpha_k\right)}{\sum_{l}^{K} \exp \left(\alpha_l\right)},
\end{equation}
as it was done in \cite{krapac2011modeling}. For each sample $\bm{x}_t$, $k$ posteriors describing the responsibility of each component for generating the sample can be found as:
\begin{equation}
\label{eq:posterior}
\gamma_t(k)=\frac{w_k u_k(\bm{x}_t)}{\sum_{l}^{K}{w_l u_l(\bm{x}_t)}}.
\end{equation}
When viewing the GMM as a neural network layer, we treat the sample $\bm{x}_t$ as the input to the layer and the posteriors $\lbrace \gamma_t(k), k = 1, ..., K\rbrace$ as its output. In the case when $\bm{x}_t$ is a $n_c$-dimensional vector, we can see that $u_{k}(\bm{x}_t)$ can be calculated by using subtraction, addition, multiplication, division and exponentiation which all are standard operations found in neural networks.

In a more general case, $\bm{x}_t$ can be treated as a $3$-dimensional tensor, that can correspond to a feature map consisting of $n_c$ channels of height $F_h'$ and width $F_w'$. This tensor could, for example, be an image with 3 RGB channels, or any feature map outputted by a preceding layer. Following the same idea of convolution in convolutional layers, we can calculate both $u_{k}(\bm{x}_t)$ and $\lbrace \gamma_t(k), k = 1, ..., K\rbrace$ by performing the same operations we did in the case when $\bm{x}_t$ was a vector, repeating them for each of the $F_h' \cdot F_w'$ $n_c$-dimensional vectors in the tensor. This procedure will result in also a $3$-dimensional tensor of size $F_h' \times F_w' \times K$. This is easily extended to the case of $4$-dimensional tensors that usually appear in deep learning frameworks, with the first dimension corresponding to the number of samples in a minibatch.

\subsection{Fisher vector descriptor layer}
\label{section:fisher}
The Fisher vector descriptor is a global descriptor used to represent data by
describing how the parameters of a generative model fitted on a distribution of
local features should change in order to better model the local features
extracted from the given data sample. An introduction to Fisher vectors and the underlying theory can be found in \cite{sanchez2013image}. In this subsection we show how the Fisher vector descriptors are calculated.
If we define the statistics of the GMM as:
\begin{equation}
\label{eq:s0}
S_k^0=\sum_{t=1}^{T}\gamma_t(k),
\end{equation}

\begin{equation}
\label{eq:s1}
\bm{S}_k^1=\sum_{t=1}^{T}\gamma_t(k)\bm{x}_t,
\end{equation}
and
\begin{equation}
\label{eq:s2}
\bm{S}_k^2=\sum_{t=1}^{T}\gamma_t(k)\bm{x}_t^2,
\end{equation}
where $\gamma_t(k)$ is the $k$-th posterior from Equation \ref{eq:posterior}, the parts of the Fisher vector corresponding to each of the parameters of the GMM can be calculated as follows:
\begin{equation}
\mathscr{G}^{X}_{w_k}=\left(S_{k}^{0} - Tw_k \right) / \sqrt{w_k},
\end{equation}
where $T$ represents the number of local descriptors,
\begin{equation}
\mathscr{G}^{X}_{\mu_k}=\left(\bm{S}_{k}^{1} - \bm{\mu}_k S_{k}^{0} \right) / \left(\sqrt{w_k}\bm{\sigma}_k\right),
\end{equation}
and
\begin{equation}
\mathscr{G}^{X}_{\sigma_k}=\left(\bm{S}_{k}^{2} - 2\bm{\mu}_k \bm{S}_{k}^{1} + (\bm{\mu}_{k}^{2} - \bm{\sigma}_{k}^{2})S_{k}^{0} \right) / \left(\sqrt{2w_k}\bm{\sigma}_{k}^{2}\right).
\end{equation}
The resulting vectors are concatenated into a large vector:
\begin{equation}
\label{eq:unnormalizedfv}
\mathscr{G}^{X}_{\lambda}=\left(\mathscr{G}^{X}_{w_1}, \ldots, \mathscr{G}^{X}_{w_K},  \mathscr{G}^{X'}_{\mu_1} \ldots,   \mathscr{G}^{X'}_{\mu_K},
\mathscr{G}^{X'}_{\sigma_1} \ldots,   \mathscr{G}^{X'}_{\sigma_K}\right)',
\end{equation}
which is the unnormalized version of the Fisher vector. The FV is normalized by applying two kinds of normalization; power normalization:
\begin{equation}
\left[\mathscr{G}_{\lambda}^{X} \right]_i \leftarrow sign\left( \left[ \mathscr{G}_{\lambda}^{X} \right]_i\right)
\sqrt{\lvert\left[ \mathscr{G}_{\lambda}^{X} \right]_i\rvert},
\end{equation}
and L2 normalization:
\begin{equation}
\label{eq:normalizedfv}
\mathscr{G}_{\lambda}^{X} \leftarrow \mathscr{G}_{\lambda}^{X} / \sqrt{\mathscr{G}_{\lambda}^{X'}\mathscr{G}_{\lambda}^{X}}.
\end{equation}

We can view the Fisher vector descriptor encoding as a network layer which contains an internal GMM layer and receives data, in the simplest case an $n_c$-dimensional vector, $\bm{x}_t$ as input. The input data is passed to the internal GMM layer which gives the posteriors $\lbrace \gamma_t(k), k = 1, \ldots, K\rbrace$ at its output. The input data and the posteriors are then used to calculate the statistics from Equations \ref{eq:s0}, \ref{eq:s1} and \ref{eq:s2}. We can see that the operations required in order to calculate both the unnormalized (Equation \ref{eq:unnormalizedfv}) and normalized (Equation \ref{eq:normalizedfv}) versions of the Fisher vector descriptor are all standard operations typically found in neural networks so the Fisher vector descriptor encoding can be easily expressed as a network layer. The dimensionality of the calculated FV is $d_{FV}=K \cdot (2n_c + 1)$.

Following the same reasoning as with the GMM layer when dealing with tensor data at input, the Fisher vector layer can also receive a tensor of size $F'_h \times F'_w \times n_c$ as its input. This tensor can be seen as a set of $F'_h \cdot F'_w$ $n_c$-dimensional vectors which we can then encode using the FV descriptor by the procedure described above. Note, that the FV encoding is performed by aggregating, across the first two modes of the tensor, the differential representations that are extracted along the fibers of the input tensor in the third mode. It is therefore trivial to extract the FV descriptor only from a selected subtensor of the input tensor. This allows us to use multiple crops of the input video during both the training and testing time in order to prevent overfitting and help improve generalization. This also allows us to create other architectures, such as the spatial pyramid \cite{lazebnik2006beyond}.


\subsection{Classification layer}
\label{section:class}

Given a Fisher vector descriptor of a video or a part of a video we design the final layer of our network to output a prediction of the input video's class. To this end, we train $m$ binary one-vs-all support vector machines as a classifier, where $m$ is the number of classes. The cost that we use for optimizing the whole network is the squared SVM hinge loss defined as:
\begin{equation}
\label{eq:svm}
C(\bm{w}, \bm{x}, \bm{y}) = \frac{\lambda}{2}||\bm{w}||^2 + \sum_j^m \max \left( 0, 1 - y_j \cdot s_j \right)^2,
\end{equation}
where $\bm{w}$ are the SVM weights, $\lambda = 2/(NC)$ with $C$ being a regularization constant, $\bm{y}$ is the label encoded as a vector where all elements are -1 except for one that is 1, marking which class the input $\bm{x}$ belongs to, and with $\bm{s}$ being the SVM score, $\bm{s} =\bm{x}\bm{w}^T+\bm{b}$. We use $y_j$ to denote the $j$-th element of the vector $\bm{y}$.

\subsection{Fisher vector network for action recognition}
\label{section:final}
The input to our network is a video represented as a stack of $L$ consecutive static frames. In order to explain the pipeline of our architecture we will first limit the length of the input video to $t$, with $t \ll L$. 
Each of the $t$ frames is passed through the local feature extraction layers which output $t$ feature maps of size $F_h\times F_w\times d$, where $d$ is the number of channels. These feature maps are then sent through the spatio-temporal pooling layer where they are pooled temporally and spatially as described in Subsection \ref{section:pooling}, resulting in a representation of size $F'_h\times F'_w\times D$. After passing this through the dimensionality reduction layer described in Subsection \ref{section:reduction} the new representation is of a lower dimensionality, $F'_h \times F'_w \times n_c$. This is then passed into the Fisher vector descriptor layer described in Subsection \ref{section:fisher} which internally uses the same input in the Gaussian mixture model layer from Subsection \ref{section:gmm} to get a tensor of posteriors of size $F'_h \times F'_w \times K$. We can treat the input tensor as a set of $F'_h \times F'_w$ descriptors with $n_c$ dimensions which also have corresponding $F'_h \times F'_w$ posteriors for each of the $K$ components of the GMM layer. These are then used to calculate the needed statistics and get the unnormalized version of the Fisher vector. The FV can then be normalized and fed into the classification layer which gives the predicted label for the $t$ frames of the original input.

For the unconstrained case when the whole video of $L$ frames is to be classified it is enough to notice that the unnormalized Fisher vector descriptor of a sequence of $2t$ frames is equal to the sum of the unnormalized FVs of the first $t$ frames and the second $t$ frames. Therefore, we can calculate the FV representation of the whole video by sliding our network along the time axis with a temporal stride of $\delta_T$ frames and summing the unnormalized FVs for each of the time segments. After normalizing the FV we can feed it into the classification layer to get the predicted label for the whole video. When several several crops are used, the resulting FV are fed to the classification layer and the corresponding outputs averaged.

The proposed neural network could also be viewed as a 3D "filter" with an internal representation which changes as the filter is "convolved" through a given video represented as a spatio-temporal volume. Once the filter passes through the whole video, the classification layer of the network uses the internal video representation, \ie the Fisher vector descriptor, to give a prediction about the video's label.

\subsection{Number of trainable parameters}
As our proposed network is a fully convolutional network, the number of its trainable parameters does not depend on the input's dimensions. Here we will summarize the total number of parameters learnt in each of our architecture's layers, excluding the local feature extraction layers as these can be replaced by an arbitrary network.

The PCA layer consists of two trainable parameters; a $D-$dimensional mean vector $\bm{\mu}_{x}$ and a $n_c \times D$ dimensional matrix of principal axes $\bm{P}$, where $n_c$ denotes the number of components kept after applying the projection and $D$ is the number of channels of the tensor returned from the spatio-temporal pooling layer.
The FV layer consists of a GMM layer that contains $K$ trainable sets of parameters $w_k$, $\mu_k$ and $\Sigma_k$, where $w_k$ is a scalar, $\bm{\mu}_k$ is a $n_c$-dimensional vector and $\bm{\Sigma}_k$ is diagonal matrix containing $n_c$ trainable parameters.
The classification layer consists of a $m\times d_{FV}$ dimensional matrix and a $m$-dimensional vector, where $m$ is the number of classes and $d_{FV}$ is the dimensionality of the FV layer output, $d_{FV} = K \cdot (2n_c + 1)$. In total, the top layers of our architecture contain $D \cdot (n_c + 1) + K \cdot (m \cdot (2n_c + 1) + 2n_c + 1) +m$ trainable parameters. As a concrete example, the top layers of the architecture we finetuned on the UCF-101 dataset (Table \ref{table:vgg16conv5}) with $n_c=100$, $K=256$, $D=6144$ and $m=101$ contained 5\,869\,157 trainable parameters.

The dimensionality of the last pooling layer in the VGG-16 \cite{simonyan2014very} architecture for a single input is (512, 7, 7). The pooling layer is fully connected to 4096 units, followed by two more fully connected layers containing 4096 and 1000 units respectively. Including the biases, this corresponds to having 123\,642\,856 trainable parameters after the convolution and pooling layers. In case of the last fully connected layer having only 101 units (when applied to the UCF-101 dataset), the parameter count is 119\,959\,653.
Similarly, the fully connected layers in the CNN-M-2048 \cite{chatfield2014return, simonyan2014two} network contain 4096, 2048 and 1000 units each, amounting to 85\,941\,224 trainable parameters in the top layers. For the case when there are 101 classes, the fully connected layers contain 84\,099\,173 trainable parameters.

By replacing the fully connected layers at the end of the network with the layers we propose, the number of trainable parameters drops to under 5\% of the original number in case of the VGG-16 network, and under 7\% in case of using the CNN-M-2048 network.

\section{Experiments and results}
The UCF-101 dataset introduced in \cite{soomro2012ucf101} consists of $13320$ video clips from $101$ different classes, divided into three pairs of train and test sets. To evaluate the performance of a method on this dataset, the average accuracy over the three splits is reported. We first run our experiments only on the first split and only evaluate the most promising approach on all three splits. 


We start by implementing the method described in \cite{palasek2016marmi} in the Lasagne/Theano framework \cite{lasagne, 2016theano} and treat it as the baseline for our experiments which we perform on the UCF-101 dataset. Training the architecture included training a single layer convolutional restricted Boltzmann machine \cite{lee2009convolutional} containing 64 filters of size $5\times5$px, learning a PCA projection ($n_c=100$), training a GMM ($K=256$) using the expectation-maximization algorithm and training a multi-class SVM classifier ($C=100$). All these steps, except for training of the SVM are done in an unsupervised manner.
After initializing the parameters of our architecture with the ones we got by the unsupervised training steps mentioned above, we did one epoch of finetuning of the whole network using the AdaGrad adaptive gradient algorithm \cite{duchi2011adaptive}. 

The size of the temporal window, \ie the number of frames needed to calculate a single FV descriptor, is set to $t=15$ in all of our experiments. The size of the spatial window in the spatio-temporal pooling layer is set to correspond to a window of $32\times32$ pixels in the input video ($S_h=S_w=7$, when the single convolutional RBM was used). These are the values used in other similar works, \eg \cite{wang2013dense}. We can control how dense we want to sample the features from the given video by setting the spatial stride parameter $\delta_S$ and the temporal stride parameter $\delta_T$. In order to decrease the time needed to do a single finetuning pass through the training set, we use $\delta_S=7$ "pixels" (corresponding to 16 pixels in the input video) and $\delta_T=15$ frames in most of our finetuning experiments.
One epoch of finetuning using these parameters on the whole training set takes around 10 hours on a Titan X GPU.

As can be seen from the results reported in Table \ref{table:convrbm}, our method using features extracted from a single layer convolutional RBM performs better than the other state of the art methods shown in Table \ref{table:state} that suffered from overfitting when trained only on the UCF-101 dataset. 
However, when more complex models are pretrained on datasets that provide larger amounts of data than the UCF-101 dataset, the performance of the simple single layer network is easily surpassed. This is not surprising as the simple network is too shallow to learn more discriminative features needed for action classification. To show how our proposed method works when the simple network is replaced with a more complex one, we choose the VGG-16 network from \cite{simonyan2014very} pretrained on the ImageNet dataset, which was also used in the two-stream network of \cite{feichtenhofer2016convolutional}.

The VGG-16 network consists of 13 convolutional layers, followed by 3 fully connected layers. We first use the outputs of the conv4\_3 layer as the input to the layers proposed in this work. Similar to the previously described experiment, we randomly extract $1000$ subvolumes from conv4\_3 layer's feature maps, corresponding to $32\times32$ px and $t=15$ spatio-temporal subvolumes in the original video. A subset of $30$ subvolumes per video is then used to learn a PCA mapping lowering their dimensionality to $n_c=100$. These are then used to train a GMM with $K=256$ components, which we use to extract Fisher vector descriptors from and finally train a SVM with $C=100$. We report the results of this experiment on all three splits of UCF-101 in table \ref{table:vgg16conv4}. We repeat the same procedure replacing the conv4\_3 layer by conv5\_3 and report the results in Table \ref{table:vgg16conv5}. As the features extracted from the conv5\_3 layer performed better than the ones from layer conv4\_3, we pick this layer for our finetuning experiments.

The larger network is more prone to overfitting so we regularize the finetuning using using dropout \cite{JMLR:v15:srivastava14a} ($p=0.9$) on the output of the local feature extraction layers. To maximize the amount of information available in the network during both training and testing we set the spatial stride $\delta_S=1$, but we keep the temporal stride fixed to $\delta_T=15$ as in the previous experiments.
We finetune the network using stochastic gradient descent with momentum (set to 0.9), showing the network one video at a time. The initial learning rate was set to 0.0001 and it was multiplied by a factor of 0.95 after each epoch. One epoch of finetuning took around 17 hours $\sim$ 30 FPS). Testing ran at a speed of around 40 FPS.

\begin{table}[htb]
\centering
\caption{UCF-101 split 1 classification accuracy, using features from a single
convolutional RBM trained only on UCF101.}
\label{table:convrbm}
\begin{tabular}{lccc}
\textbf{Method}     & \textbf{Split 1}          \\
\hline
Single CRBM + FV \cite{palasek2016marmi} & 55.06\% \\
\hline
Ours, single CRBM, random sampling   & 59.95\%\\
Ours, single CRBM, dense sampling  & 60.37\% \\
\hline

\textbf{Our finetuned network} (after 1 epoch) & 61.67\%\\
\end{tabular}
\end{table}

\begin{table*}[htb]
\centering
\caption{UCF-101 classification accuracy, using the conv4\_3 layer features from the VGG-16 network pretrained on ImageNet.}
\label{table:vgg16conv4}
\begin{tabular}{lcccc}
\textbf{Method}    & \textbf{Split 1}  & \textbf{Split 2} & \textbf{Split 3} & \textbf{Average}\\
\hline
Random sampling  & 70.84\% & 70.30\% & 70.81\% & 70.65\% \\
\end{tabular}
\end{table*}

\begin{table*}[htb]
\centering
\caption{UCF-101 classification accuracy, using the conv5\_3 layer features from the VGG-16 network pretrained on ImageNet. Finetuning was done using SGD with initial learning rate = 0.0001, momentum = 0.9 and extraction
layer dropout p = 0.9.}
\label{table:vgg16conv5}
\begin{tabular}{lcccc}
\textbf{Method}    & \textbf{Split 1}  & \textbf{Split 2} & \textbf{Split 3} & \textbf{Average}\\
\hline
Random sampling  & 75.65\% & 76.06\% & 74.89\% & 75.54\%\\
Dense sampling  & 75.55\% & 76.33\% & 74.35\% & 75.41\%\\
\hline
Finetuned network (after 5 epochs) & 76.39\% & 77.29\% & 77.30\% & 76.99\% \\
Finetuned network (after 11 epochs) & 79.12\% & 78.63\% & 76.35\% & 78.03\% \\ 
Finetuned network (after 33 epochs) & 81.84\% & -& - & - \\
\end{tabular}
\end{table*}

\begin{table*}[h!t!]
\centering
\caption{State of the art methods using only static features, trained and evaluated on UCF-101.}
\label{table:state}
\begin{tabular}{lccr}
\textbf{Method}    & \textbf{Split} & \textbf{Accuracy} & \textbf{Total parameters}\\
\hline
Slow fusion network \cite{karpathy2014large}  & all & 41.3\% & \\
Spatial CNN-M-2048 \cite{simonyan2014two} & 1 & 52.3\% & $\sim$ 90.63 M \\ 
Single CRBM + FV \cite{palasek2016marmi}  & 1 & 55.06\% & $\sim$ 5.33 M \\
\textbf{Ours}, single CRBM & 1 & \textbf{61.67}\%  & \textbf{$\sim$ 5.33 M} \\ 
\end{tabular}
\end{table*}

\begin{table*}[h!t!]
\centering
\caption{State of the art methods using only static features, pretrained on a larger dataset, finetuned and evaluated on UCF-101.}
\label{table:state2}
\begin{tabular}{lcccrr}
\textbf{Method}   & \textbf{Pretrained on} & \textbf{Split} & \textbf{Accuracy} & \textbf{Top layers parameters} & \textbf{Total parameters}
\\
\hline

Slow fusion network \cite{karpathy2014large} & Sports 1M & all & 65.4\% & - & - \\
Encoding objects \cite{jain201515} & ImageNet & all & 65.6\% & - & - \\
Spatial CNN-M-2048 \cite{simonyan2014two} & ImageNet & 1 & 72.8\% & 84.1 M & 90.63 M \\
\textbf{Ours}, VGG-16 & ImageNet & 1 & \textbf{81.84}\% & \textbf{5.87 M} & \textbf{20.58 M} \\
Spatial VGG-16 \cite{feichtenhofer2016convolutional,simonyan2014very} & ImageNet & 1 & 82.61\% & 119.96 M & 134.67 M \\
\end{tabular}
\end{table*}

\section{Discussion}

Compared to the work of \cite{palasek2016marmi}, where the same kind of features and encoding were used and the extraction was performed on randomly selected subvolumes, our network is naturally capable of performing dense sampling, thus increasing the available information from the underlying video and improving the final classification performance. By performing dense sampling and finetuning the whole network an improvement to 61.3\% was achieved. This is explained by the fact that including more training data helps prevent overfitting.

Let us note that the lowest level of our network extracts features at frame level from intensity information alone, and therefore is not directly comparable to the full two stream network from \cite{simonyan2014two}, one stream of which is trained on optical flow that was extracted offline at the input. However, we do compare favorably with the spatial stream of the network when trained directly on static frames of the UCF-101 dataset, where it overfits and results in a classification accuracy of 52.3\% - compared to 61.67\% that we obtain when using a simple, single-layer convolutional RBM. In order to prevent overfitting, the two stream network is pretrained on a different, larger dataset - this alone improves the accuracy of \cite{simonyan2014two} to 72.8\%. 

Our approach, which includes the time dimension by pooling feature maps extracted from subvolumes of the video, achieves an accuracy of 61.67\%, without including optical flow explicitly and only using a single convolutional RBM at the lowest layers of the network. The work of \cite{karpathy2014large}  also tried to tackle the problem of including motion features implicitly by trying to learn them from stacks of static frames. The approach of slowly fusing the feature maps resulted in an accuracy of  65.4\% on the UCF-101 dataset when pretrained on a larger (Sports 1M) dataset. Whereas training directly on UCF-101 resulted in overfitting with an accuracy of 41.3\%. This is again comparable to the 61.67\% that we obtain with the proposed approach, when using the simple single-layer convolutional RBM features.

By simply replacing the single convolutional RBM layer at the lowest level of our architecture with VGG-16, a deep network containing 13 convolutional layers pretrained on ImageNet, we boost the classification accuracy to 75.41\%. The main contribution of our proposed method is shown after performing the finetuning of the network as a whole, which further boosts the classification accuracy on UCF-101 to 81.84\%. While this is lower than the 82.61\% achieved by the VGG-16 network as the spatial stream of \cite{feichtenhofer2016convolutional}, we point out that our architecture contains 20.58 million trainable parameters in total, compared to the 134.67 million parameters contained in VGG-16. If we only look at the top layers of the two architectures, the 3 fully connected layers containing 119.96 million parameters in VGG-16 can be replaced by our proposed layers that contain only 5.87 million parameters, that is less than 5\% of the parameter count, at the cost of diminishing the classification performance by 3.5\%. Longer finetuning and a finer choice of the finetuning hyperparameters should lower this performance gap. On the other hand, our method compares favorably to the CNN-M-2048 spatial stream of \cite{simonyan2014two}, achieving 81.84\% versus its 72.8\%, while requiring less than 23\% of its total trainable parameter count.

\section{Conclusion}

In this paper we have proposed a convolutional architecture that expresses the various steps of the Fisher vector based action recognition as layers in convolutional neural network that can be trained or refined end to end in a supervised manner. 
Our model outperforms significantly the baseline architecture where the various levels are trained in a layer by layer manner unsupervised, and state of the art CNN architectures when trained on the same amount of data. We show that replacing the top fully connected layers in popular convolutional network architectures with our proposed layers results in a significant reduction of the needed trainable parameter count, while achieving a comparable performance, or even significantly surpassing the performance of similar architectures.

{\small
\bibliographystyle{ieee}
\bibliography{bibliography}
}

\end{document}